\documentclass[runningheads]{llncs}
\usepackage{graphicx}
\usepackage{comment}
\usepackage{siunitx}
\usepackage{booktabs}
\usepackage{xcolor}
\usepackage{hyperref}
\usepackage{bm}
\usepackage{amsmath}
\usepackage{amsfonts}
\usepackage{soul}

\hypersetup{
    colorlinks,
    linkcolor={blue!50!black},
    citecolor={blue!50!black},
    urlcolor={blue!50!black}
}

\begin{document}
\title{Multiple Instance Learning with Auxiliary Task Weighting for Multiple Myeloma Classification}
\titlerunning{Multiple Instance Learning with Auxiliary Task Weighting}
\author{Talha Qaiser \inst{1} \and
Stefan Winzeck\inst{1} \and
Theodore Barfoot \inst{1,2} \and 
Tara Barwick \inst{3,4} \and
Simon J. Doran \inst{5} \and
Martin F. Kaiser \inst{5,6} \and
Linda Wedlake \inst{6} \and
Nina Tunariu \inst{6} \and
Dow-Mu Koh \inst{5,6} \and
Christina Messiou \inst{5,6} \and
Andrea Rockall \inst{3,4} \and
Ben Glocker \inst{1}}

% index{Qaiser, Talha}
% index{Winzeck, Stefan}
% index{Barfoot, Theo}
% index{Barwick, Tara}
% index{Doran, Simon J.}
% index{Kaiser, Martin F.}
% index{Wedlake, Linda}
% index{Tunariu, Nina}
% index{Koh, Dow-Mu}
% index{Messiou, Christina}
% index{Rockall, Andrea}
% index{Glocker, Ben}

%
\authorrunning{Qaiser et al.}
\institute{BioMedIA Group, Department of Computing, Imperial College London, UK \and
Biomedical Engineering and Imaging Sciences, King’s College London, UK \and
Department of Surgery and Cancer, Imperial College London, UK \and 
Department of Imaging, Imperial College Healthcare NHS Trust, London, UK \and
The Institute of Cancer Research, London, UK \and
The Royal Marsden NHS Foundation Trust, UK \\
\email{\{t.qaiser,b.glocker\}@imperial.ac.uk}}
\maketitle              % typeset the header of the contribution
\begin{abstract}
Whole body magnetic resonance imaging (WB-MRI) is the recommended modality for diagnosis of multiple myeloma (MM). WB-MRI is used to detect sites of disease across the entire skeletal system, but it requires significant expertise and is time-consuming to report due to the great number of images. To aid radiological reading, we propose an auxiliary task-based multiple instance learning approach (ATMIL) for MM classification with the ability to localize sites of disease. This approach is appealing as it only requires patient-level annotations where an attention mechanism is used to identify local regions with active disease. We borrow ideas from multi-task learning and define an auxiliary task with adaptive reweighting to support and improve learning efficiency in the presence of data scarcity. We validate our approach on both synthetic and real multi-center clinical data. We show that the MIL attention module provides a mechanism to localize bone regions while the adaptive reweighting of the auxiliary task considerably improves the performance. 

\end{abstract}
\section{Introduction}
Caused by the aberrant proliferation of plasma cells within the bone marrow, multiple myeloma (MM) is one of the most common hematologic malignancies. While MM is incurable, it can be treated when diagnosed appropriately \cite{Kumar2017}. Diffusion-weighted (DW) WB-MRI offers high sensitivity to diagnose MM in an early stage \cite{messiou2018whole}. DW WB-MRI is a relatively new diagnostic tool, only recently being recommended for patients with myeloma but not yet widely used. To our knowledge, there are currently no computational approaches for MM classification using DW WB-MRI. Prior work in this domain mostly focused on PET/CT MM imaging \cite{mesguich2021improved,xu2018automated}. Broadly, bone marrow appearances of patients imaged for suspected myeloma can be classified as: healthy, focal lesions, diffuse infiltration, or focal lesions on a background of diffuse infiltration. The presence of more than one focal lesion with diameter $> 5 \si{\milli\metre}$ indicates the need for treatment, smaller lesions will require monitoring \cite{messiou2019guidelines}. In clinical practice, myeloma patterns are identified by visual examination of WB-MRI scans. However, this compares to the search of a needle in a haystack as WB-MRI consists of hundreds of 2D images and lesions can be scattered across the skeletal system and make up only a small fraction ($\approx \leq10\%$). Automated MM classification with the ability to localize disease could improve diagnostic accuracy and reduce the reading time.

\begin{figure}[t]
\centering
\includegraphics[width=0.85\textwidth]{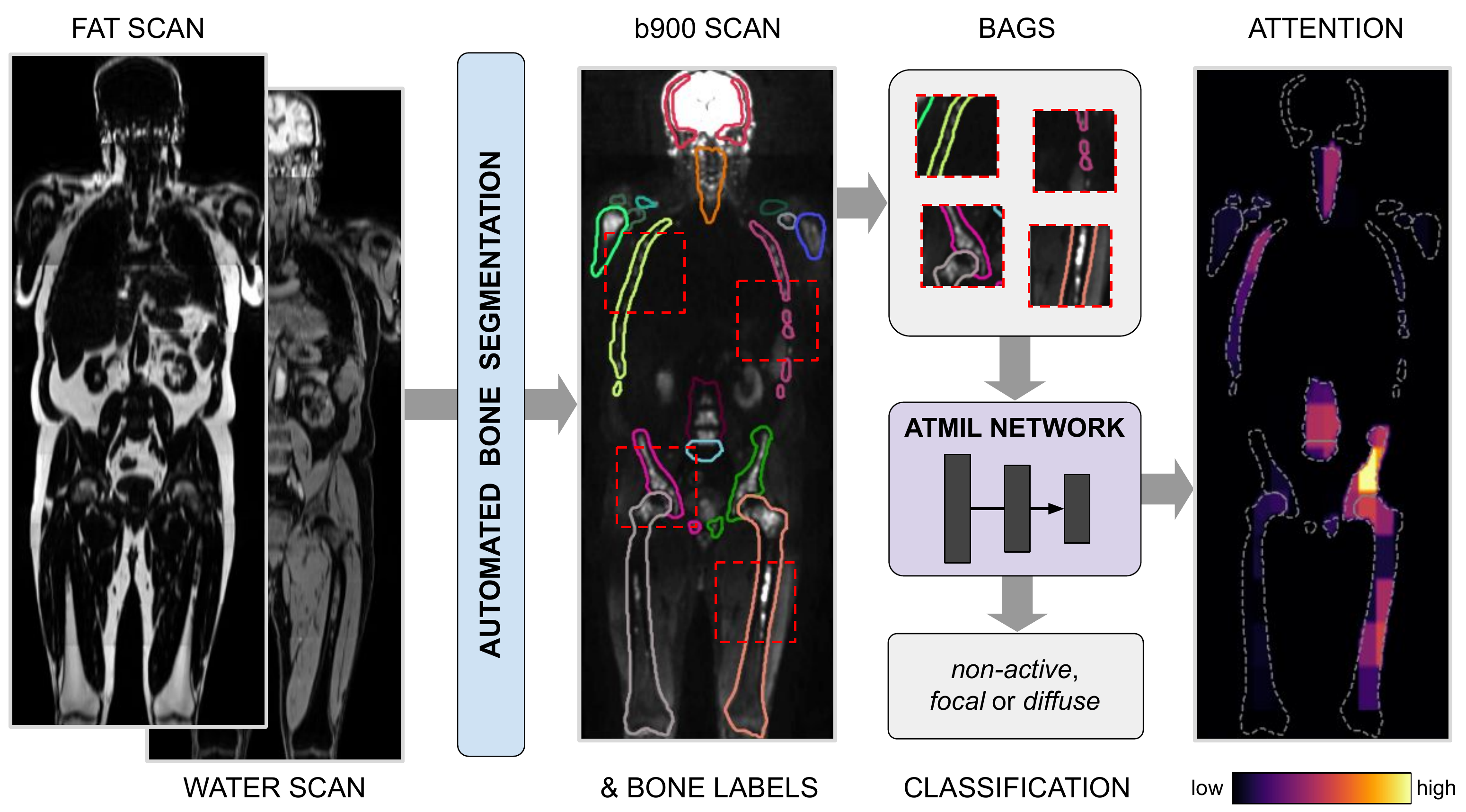}
\caption{Schematic overview of the proposed framework. Fat and water WB-MRI scans are used to automatically segment bone structures. For each b900 scan, 3D instances from segmented bone regions are sampled and combined into a bag. From these, the ATMIL network derives a disease class and an attention map.} \label{fig1}
\end{figure}

Convolutional neural networks (CNNs) are successfully applied for whole-image disease classification, but lack the ability to localize disease. To train a CNN for local disease classification one would require region-wise or voxel-wise annotations, which are difficult to obtain in large quantities. Weakly supervised approaches, such as multiple instance learning (MIL), are appealing as these can be trained on global, patient-level labels (e.g. disease categories) while having the ability to localize the contributing image regions. Instead of predicting labels for individual samples (e.g. patches), MIL frameworks aim to derive a single label from a set of instances, known as a \textit{bag}. Broadly, deep MIL can be categorized into two groups (a) multi-stage, where an encoder maps bag instances into a low-dimensional feature space and an aggregation model is then used to predict bag labels \cite{campanella2019clinical,ozdemir20193d,lu2020data}, (b) end-to-end, where both the low-dimensional embedding and bag label predictions are learned by a single model \cite{ilse2018attention,fernando2019deep,chikontwe2020multiple}. End-to-end models equipped with attention mechanisms offer better interpretability and localisation ability and have recently been shown to outperform multi-stage models \cite{chikontwe2020multiple}. MIL has been successfully applied to large data sets, mostly in histopathology, but have not been studied for WB-MRI disease classification where training data is scarce and patterns of disease can be subtle.

\subsubsection{Contributions.} In this work, we propose an ATMIL network to classify MM disease patterns in 3D WB-MRI. To localize disease, a permutation-invariant attention mechanism assigns higher weights to bone structures that are more likely to carry MM lesions without the need for any localized annotations. To combat data scarcity and enable the model to generalize well, we borrow ideas from multi-task learning (MTL) where auxiliary tasks are designed to support learning of the main task \cite{lin2019adaptive,shi2020auxiliary}. We propose adaptive reweighting by minimizing the divergence between our main and auxiliary task. We have compared the performance of our approach in combination with different auxiliary losses. At first, we evaluated our framework's performance on Morpho-MNIST \cite{castro2019morpho}, a synthetic data set we use to mimic a 4-class classification problem. We then validate the performance on real, multi-center clinical data with diffusion weighted WB-MRI. Our results suggest that the auxiliary task improves the generalization and predictive performance of the attention-based MIL framework with the ability to identify anatomical regions containing local disease patterns.

\section{Methodology}
Our proposed framework contains three main components  (Fig.\ref{fig1}): an automated bone segmentation to identify regions of interests for sampling bag instances, an ATMIL network for MM classification with an attention mechanism to localize disease, and an integrated adaptive weighting scheme to quantify the usefulness of an auxiliary task to support learning of the main task.

\subsection{Multiple Instance Learning}
\begin{figure}[t]
\centering
\includegraphics[width=1.00\textwidth]{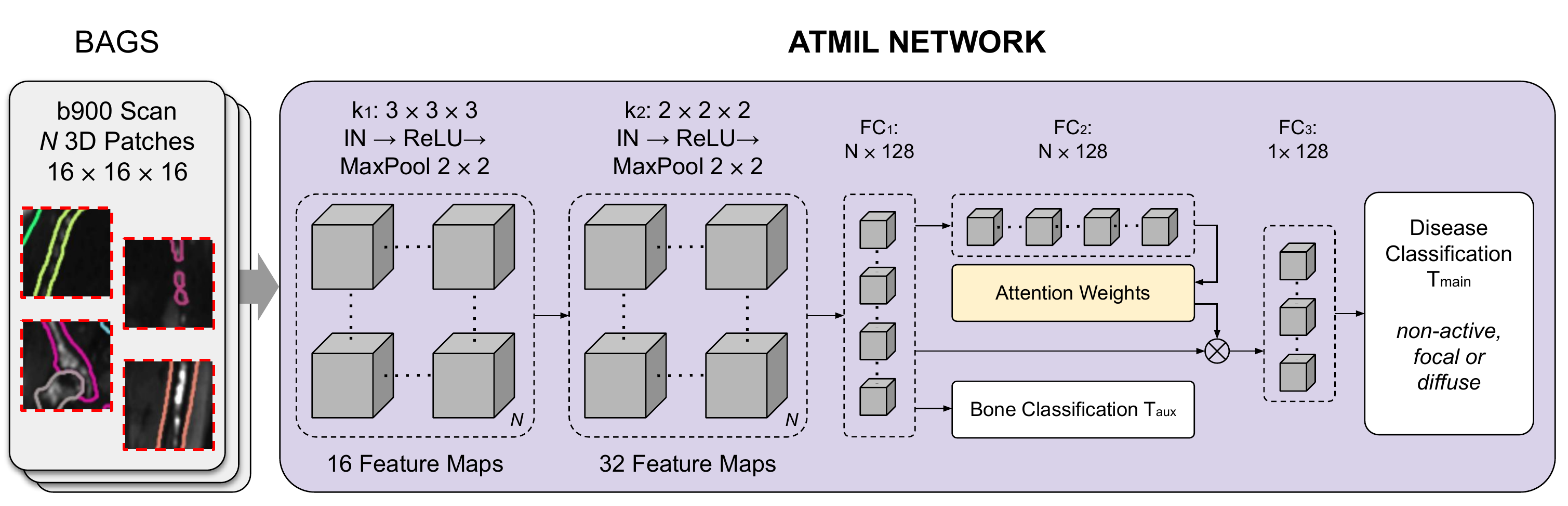}
\caption{Illustration of the proposed ATMIL framework architecture. Bags of 3D instances (from segmented bone regions) are passed through convolutional (k1 \& k2) and fully connected layers (fc1, fc2 \& fc3). The network provides a disease classification ($T_{main}$), an auxiliary bone classification ($T_{aux}$) and attention maps.} \label{fig2}
\end{figure}

In a fully-supervised learning problem the objective is to predict from images $I\in\mathbb{R}^{D}$ a target class label $Y\in\{1, ..., y\}$. In a MIL setting, each $I$ is represented by a bag of instances, $\delta = \{i_1, ..., i_N\}$ which is associated with only a single label $Y$. A classifier $g_\theta(\cdot)$ processes the entire bag of instances to predict the bag label. In our case, each scan represents one bag ($\delta$) and we sample 3D instances only from segmented bone regions. Figure \ref{fig2} illustrates the architecture of the proposed framework. First, a group of convolution and fully connected layers transform $i_N$  into a non-linear feature representation $\bm{h}_N$. Further, we need a symmetric function $S(\cdot)$ to aggregate instance-level embeddings $\bm{h}_N$ and to enable permutation invariance within the model:
\begin{equation} \label{eq:1}
g_\theta(I) =  S(f(i_1), ...., f(i_N))
\end{equation}
where $\bm{h}_N = f(i_N)$. For that, pooling operators (e.g. mean, max \textit{etc.}) could be applied, but standard methods are non-trainable and may hamper identifying key instances within a bag.  

\subsection{Attention-Based Multiple Instance Learning}
A soft-attention mechanism offers an adaptive and reliable alternative to pooling operators and which can also be trained in an end-to-end manner. The objective here is to use instance embeddings $H = \{h_1, ..., h_N\}$ and learn a set of attention weights $A = \{a_1, ..., a_N \}$. Further, we compute the weighted sum of instance embeddings and attention weights to obtain bag-level predictions:
\begin{equation}
z = \sum_{n=1}^{N} {a_n}{\bm{h}_n} 
\quad\quad\quad \text{with} \quad\quad\quad
a_n =  \dfrac{\exp \{ \bm{u}^T \tanh(\bm{V}\bm{h}_n^{T}) \} } {\sum_{j=1}^{N} \exp\{ \bm{u}^T \tanh(\bm{V}\bm{h}_j^T) \} }
\end{equation}
where $\bm{u}$ and $\bm{V}$ are model parameters. The attention weights also represent the importance of each instance in a bag, which we later use to compute attention maps on the WB-MR image level (as further discussed in the Results section).

\subsection{Adaptive Reweighting of Auxiliary Task}
Transferring knowledge from a related auxiliary task can provide additional supervision for the main task using the same data. It has been previously observed that multi-task learning with shared feature representation offers improved generalization \cite{girshick2015fast} and overcome data inefficiency in both supervised and reinforcement learning tasks \cite{lin2019adaptive}. The most straightforward way for optimizing MTL parameters is to combine loss functions uniformly or use a predefined hyperparameter coupled with some annealing factor (e.g. based on epochs) to reduce the influence of auxiliary task as the training proceeds \cite{sadafi2020attention}. However, while optimizing the weights for MTL, it is not obvious at what stage of the training process and how much an auxiliary task may assist the main task. Hence, quantifying the benefit of an auxiliary task is highly relevant for MIL.

Our main task ($T_{main}$) is to predict disease labels on image-level and we employ an auxiliary task ($T_{aux}$) to predict anatomical labels for bone regions (obtained from the predicted segmentation maps). The corresponding loss terms for both tasks are $\mathcal{L}_{main}$ and $\mathcal{L}_{aux}$, respectively. The proposed framework has a shared backbone with learning parameters $\theta$ and a separate head for each task. The objective is to jointly optimize model weights for the main task while leveraging the auxiliary task for additional supervision to learn more generalized features. Our framework's total loss is formalized as:  
\begin{equation}
\mathcal{L}(\theta_t) = \mathcal{L}_{main} (\theta_t) + \sum\limits_{m=1}^{M} w_m \mathcal{L}_{aux,m} (\theta_t)
\end{equation}
\begin{equation}
\theta_{t+1} =  \theta_t -  \alpha_t \nabla_{\theta_t} \mathcal{L}(\theta_t)
\end{equation}
where $w_m$ represents the weight for $m$ auxiliary task and $\theta_t$ denotes the learning parameters of the model during training step $t$. The weight update of the model parameters on this combined loss term is then:
\begin{equation}
\theta_{t} \gets \theta_{t-1} - \alpha_t (- \nabla \log p(\textit{T}_{main}| \theta_{t-1}) - \sum\limits_{m=1}^{M} w_m \nabla \log p({\textit{T}_{aux,m}} | \theta_{t-1}))  
\end{equation}
One way of estimating $w_m$ is to minimize the Fisher divergence between gradients of our $T_{main}$ and $T_{aux}$. This has been reported to be beneficial for semi- and unsupervised learning \cite{shi2020auxiliary} and offers better divergence than Kullback-Leibler \cite{yang2019variational}. Higher values of $w_m$ correspond to $T_{aux}$ which are more advantageous for $T_{main}$ or where parameter distributions of both tasks are similar. 
\begin{equation}
w \gets w - \beta \nabla_w \vert \vert \nabla \log p(\textit{T}_{main}|\theta_t ) -  \sum\limits_{m=1}^{M} w_m \nabla \log p({\textit{T}_{aux,m}} | \theta_t)  \vert \vert_2^2  
\end{equation}
with learning rate $\beta$ for the adaptive task weight. For MM disease classification, we are using a single auxiliary task (bone labels) by setting $M=1$.

\section{Data \& Implementation Details}
\subsubsection{Morpho-MNIST (proof-of-concept).} 
\begin{figure}[t]
\centering
\includegraphics[width=1.00\textwidth]{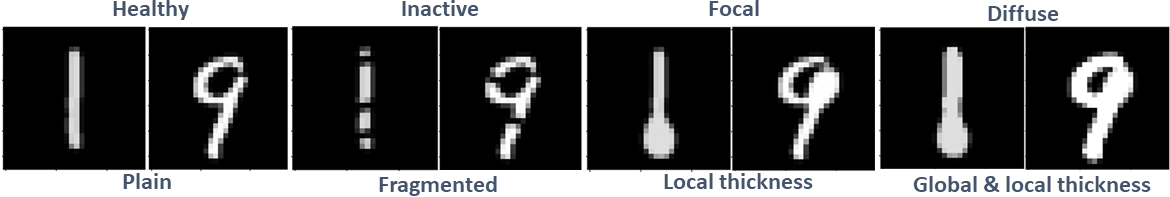}
\caption{Examples from the Morpho-MNIST Data Set. Digits were transformed to fragments, local thickness and global \& local thickness to imitate inactive, focal and diffuse MM classes, respectively.} \label{fig3}
\end{figure}

The Morpho-MNIST \cite{castro2019morpho} data set contains morphologically perturbed digits. With randomly varying intensity and thickness transformation we aim to mimic MM disease patterns  (Fig. \ref{fig3}). These include plain digits with no transformations (healthy), fragmented digits that are split into several components (inactive), local thickness (focal) as well as combined global and local thickness (diffuse). 

\subsubsection{WB-MRI (real clinical data).} 
Images were collected at The Royal Marsden Hospital and Imperial College Healthcare NHS Trust either on a Siemens MAGNETOM Aera or Avanto scanner. Structural (Dixon: fat \& water images) and diffusion (b=$50,900$ s$/\si{\milli\metre}^2$) WB-MRI data were acquired for 176 MM patients at multiple time points. The total data set of 304 scans comprises 25 healthy, 52 inactive, 110 focal as well as 117 diffuse lesion scans. The healthy and inactive classes were coined `non-active', and focal and diffuse classes refer to `active' disease patterns. Our analysis made use of 75 manually annotated fat and water scans for bone segmentation and focused on b900 WB-MRI for disease classification, as higher b-values highlight MM \cite{messiou2018whole,messiou2015whole}. All images were resampled to $2\,\si{\milli\metre}$ isotropic voxel space and standardized via Nyul's histogram matching \cite{nyul1999standardizing} to account for scanner differences.

\subsubsection{Bone Segmentation.} DeepMedic \cite{kamnitsas2017efficient} was trained and validated on 55 and 20 subjects, respectively, to segment 18 bone structures from fat and water scans (three pathways, RMSprop, learning rate = 0.001, momentum = 0.6, patch size = 37$\times$37$\times$37 voxels, L1 and L2 regularization, standard scaling of input images). To remove spurious predictions, segmentations were corrected via comparison to an atlas based segmentation, positioning to the midsagittal plane and morphological operators. Final segmentations on validation data showed average Dice scores per bone class between 0.690 (sacrum) and 0.919 (left femur). A new model was trained on all 75 annotated scans and applied to all 304 scans. Segmentations were again corrected as described above and dilated to be more inclusive.

\subsubsection{Disease Classification.} 
We sampled 3D patches of size $16\times16\times16$ that overlapped with the bone segmentation by 50\% , yielding on average $\approx755$ instances per WB MR image.
The  ATMIL network was trained with Adam optimizer \cite{kingma2014adam} with learning rate $\alpha = 0.0005$, a decaying factor of 10 (applied after 100 epochs), and adaptive task weight $\beta= 0.05$.
We trained the ATMIL network for 200 and 1000 epochs for Morpho-MNIST and WB-MRI, respectively. Random data augmentations
(cropping, rotating, horizontally flipping, elastically deformation and Gaussian smoothing $(0.0 < \sigma \leq 1.0)$) was applied to WB-MRI. Further details can be found online: \url{https://github.com/biomedia-mira/atmil}

\section{Results}
\begin{table}[t]
\setlength{\tabcolsep}{10pt}
\centering
\caption{Classification accuracy on Morpho-MNIST data. Each bag contains 100 instances and the test data consists of 1000 bags with equal representation of positive (focal, diffuse, healthy, inactive) and negative (healthy, inactive) bags.}
\begin{tabular}[t]{lccccc}
\toprule
Number of training bags & 100 &150 & 200 & 300 & 500\\
\midrule
MIL + max &0.778&0.782&0.826&0.91&0.947\\
MIL + att &0.805&0.786&0.857&0.975&0.981\\
\midrule
MIL + att + uniform &0.863&0.923&0.936&\textbf{0.985}&0.983\\
MIL + att + WL\cite{sadafi2020attention} &0.817&0.940&0.943&0.978&0.974\\
MIL + att + GN\cite{chen2018gradnorm} &0.844&0.916&0.968&0.969&0.985\\
MIL + att + CS\cite{du2018adapting} &0.824&0.939&0.964&0.983&0.985\\
MIL + att + AL\cite{hu2019learning} &0.973&0.923&\textbf{0.981}&0.975&0.986\\
MIL + att + OA\cite{lin2019adaptive} &0.931&0.953&0.963&0.982&0.987\\
ATMIL (ours) &\textbf{0.975}&\textbf{0.960}&0.951&0.976&\textbf{0.989}\\
\bottomrule
\end{tabular}
\end{table}%

\subsubsection{Comparative Analysis.}
The main objectives for reporting this analysis is (a) to investigate how the proposed framework performs with and without attention mechanism on a controlled (Morpho-MNIST) and a clinical MM (WB-MRI) dataset, (b) how different task reweighting approaches could provide gradient directions that yield additional supervision to the main task. For comparative analysis, we used the following auxiliary task reweighting methods (I) uniform (same weights for both losses), (II) weighted loss (WL) \cite{sadafi2020attention}, $(1-\gamma^{\eta}) \nabla \log p(T_{main}|\theta)$ +  $\gamma^{\eta} \nabla \log p(T_{aux}|\theta)$ where $\gamma$ represents a predefined weights $(\gamma = 0.5)$ and $\eta$ denotes number of epochs,  (III) GradNorm (GN) \cite{chen2018gradnorm} to balance both tasks’ gradient norm, (IV) AdaLoss (AL) \cite{hu2019learning} performs log operator around the auxiliary task, (V) CosineSim (CS) \cite{du2018adapting}, the auxiliary task only contribute if having positive cosine similarity, $ \cos(\nabla \log p(T_{main}|\theta), \nabla \log p(T_{aux}|\theta))$ (VI) OL-AUX (OA) \cite{lin2019adaptive} assigns high weight based on inner product of both tasks, $\nabla \log p(T_{main}|\theta) \nabla \log p(T_{aux}|\theta)^T$.
\begin{table}[t]
\footnotesize
\setlength{\tabcolsep}{8pt}
\centering
\caption{Classification results on 3D WB-MRI of MM patients. Experiments were performed 3 times and average $\pm$ standard error of the mean for specificity, sensitivity, and F1-score (macro-average) are reported.}
\begin{tabular}[t]{lcccc}
\toprule
Method & Specificity  & Sensitivity & F1-score\\
\midrule
MIL + max & 0.753 $\pm$ 0.023 &0.527 $\pm$ 0.023&0.539 $\pm$ 0.012\\
MIL + att &0.844 $\pm$ 0.013 &0.689 $\pm$ 0.027&0.702 $\pm$ 0.027 \\
\midrule
MIL + att + uniform &0.836 $\pm$ 0.010&0.672 $\pm$ 0.020 &0.689 $\pm$ 0.021\\
MIL + att + WL \cite{sadafi2020attention} &0.839 $\pm$ 0.010 &0.678 $\pm$ 0.019 &0.686 $\pm$ 0.018 \\
MIL + att + GN \cite{chen2018gradnorm} &0.861 $\pm$ 0.010 &0.722 $\pm$ 0.014&0.727 $\pm$ 0.019 \\
MIL + att + CS \cite{du2018adapting} &0.842 $\pm$ 0.010 &0.683 $\pm$ 0.095&0.712 $\pm$ 0.010\\
MIL + att + AL \cite{hu2019learning} &0.833 $\pm$ 0.010 &0.667 $\pm$ 0.018&0.671 $\pm$ 0.017\\
MIL + att + OA \cite{lin2019adaptive} & 0.850 $\pm$ 0.010 &0.701 $\pm$ 0.019 &0.702 $\pm$ 0.018 \\
ATMIL (ours) &\textbf{0.869 $\pm$ 0.010 }&\textbf{0.737 $\pm$ 0.016}&\textbf{0.744 $\pm$ 0.019 }\\
\bottomrule
\end{tabular}
\end{table}%
We perform three repetitions per experiments for the clinical MM data and report the average performance and standard error of the mean. Table 1 reports the disease classification accuracy on Morpho-MNIST data (the auxiliary task was to learn if a given image contains more than one connected component or not) and, Table 2 presents results for 3D WB-MRI when comparing with different task reweighting approaches. For a fair analysis, we randomly split our data on patient-level by selecting around 75\% for training, 5\% for validation, and 20\% for the final testing. The test data contains equal number of samples from each class (20 WB-MR scans/class). In the ATMIL setting, we are only concerned with the performance of our main task. Evidently, auxiliary task reweighting offers better generalization as compared to the uniform summation of loss terms (uniform) and other weighting methods. CosineSim \cite{du2018adapting} acts as a filter that only combines both loss terms when they follow the same direction and AdaLoss \cite{hu2019learning} simply wraps the auxiliary loss term with a log operator. Performance is comparable on Morpho-MNIST, but less good on clinical data. GradNorm \cite{chen2018gradnorm} is a gradient norm balancer and shows relatively stable performance. The proposed ATMIL approach outperforms other methods on both data sets, and most importantly on the real clinical application.

\begin{figure}[]
\centering
\includegraphics[width=1.00\textwidth]{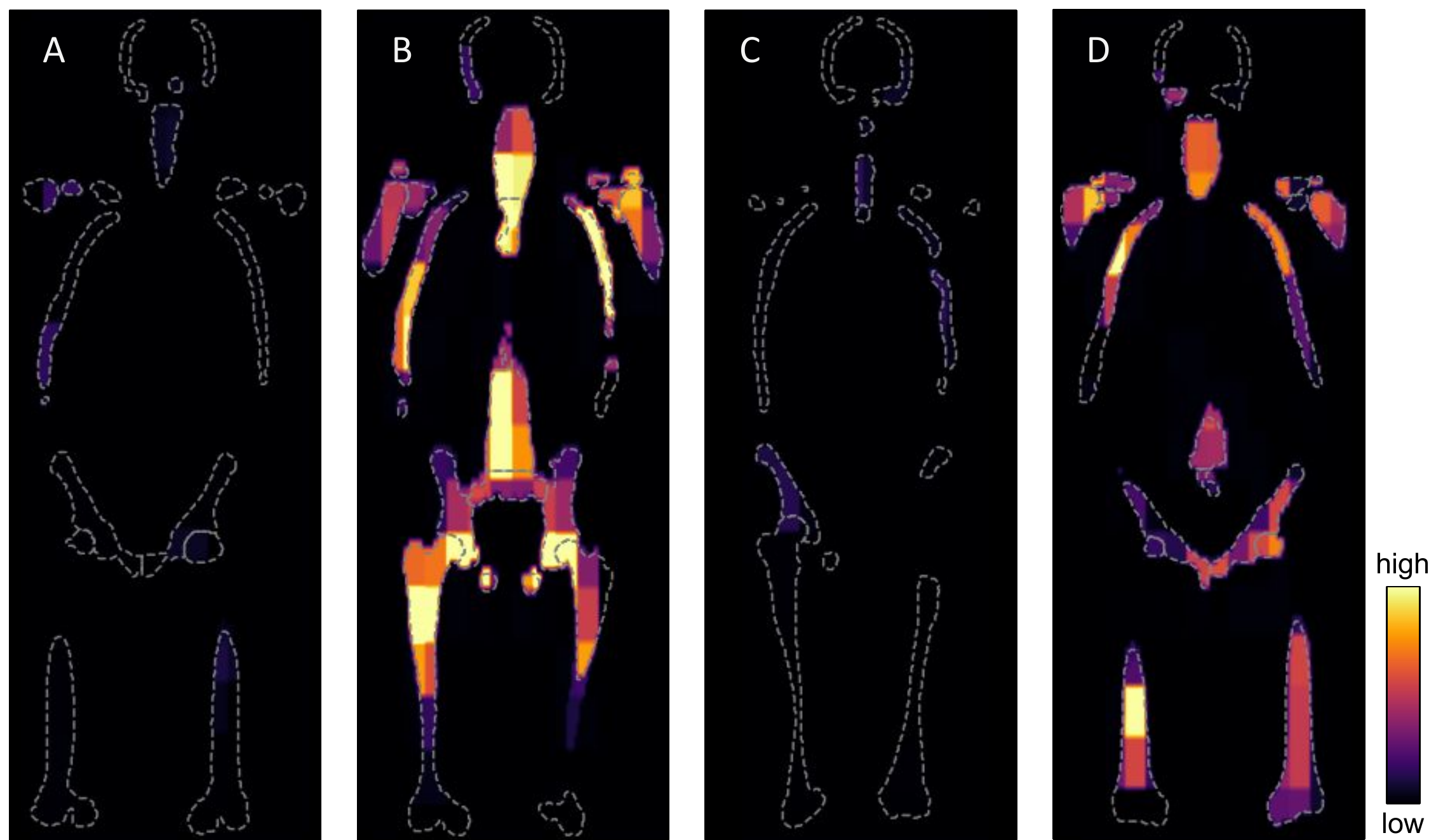}
\caption{Attention maps generated by our ATMIL network. Examples of patients with different MM reports: low (A) and high (B) occurrence of diffuse lesions \& low (C) and high (D) count of bones with focal lesions. Grey outlines show segmented bones.
} \label{fig4}
\end{figure}
\subsubsection{Qualitative Assessment.}
Figure \ref{fig4} shows qualitative, visual results for the attention maps generated by our approach. Anatomical regions that are relevant for triggering the bag label prediction are assigned high weights. We show four example patients for which different disease patterns were reported during clinical assessment. The attention maps seem to reflect well how much a scan was affected by MM. It is important to highlight that our framework was trained entirely with patient-level disease labels, and yet, is capable of highlighting burden of disease. The clinical utility of attention maps remains to be evaluated, but they may provide an interpretable output for whole-image disease classification, and could potentially even directly support radiological reading when used as a visual guide. This could be particularly useful for less experienced readers, and may generally reduce reading time of WB-MRI.
 
\section{Conclusion}
In this work, we proposed an MIL framework that leverages additional supervision from an auxiliary task using the same data. The presented method provides an effective way for adjusting the weight (to measure the usefulness) of the auxiliary task, minimizing the divergence between auxiliary and main tasks in an MIL setting. In future work it may be worth considering multiple related auxiliary tasks. As a real-world use case, we report results on WB-MRI of MM patients for which only patient-level disease labels were used during training. Attention weights enable the model to identify and localize instances corresponding to bone regions that likely show signs of disease. We believe the proposed approach carries the potential to solve other problems where only image-level labels are available and where data scarcity is a major challenge for effective training.

\section*{Acknowledgements}

This work is supported by the National Institute of Health Research (EME Project: 16/68/34). SW is funded by the UKRI London Medical Imaging \& Artificial Intelligence Centre for Value Based Healthcare. TB and AR are supported by the Imperial NIHR BRC and the Imperial CRUK Centre.

\bibliographystyle{splncs04}
\bibliography{refs}

\newpage
\appendix

\section{Multiple Myeloma Confusion Matrices}
\begin{figure}[h]
\centering
\includegraphics[width=1.0\textwidth]{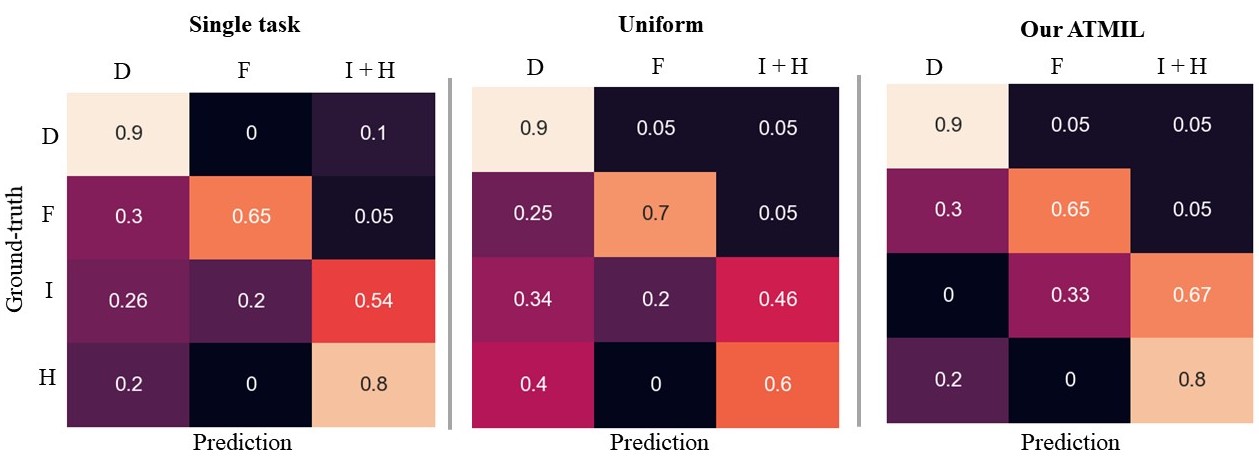}
\caption{Confusion matrices of the single task MIL (left), uniformly combining loss terms (center) and the proposed method (right) on the test data set. Each column of the matrix represents the percentage of instances in a predicted class, and each row represents the percentage of instances in an actual class (ground-truth).} \label{Sfig1}
\end{figure}
\newpage
\section{MorphoMNIST Qualitative Results}
\begin{figure}[h]
\centering
\includegraphics[width=1.00\textwidth]{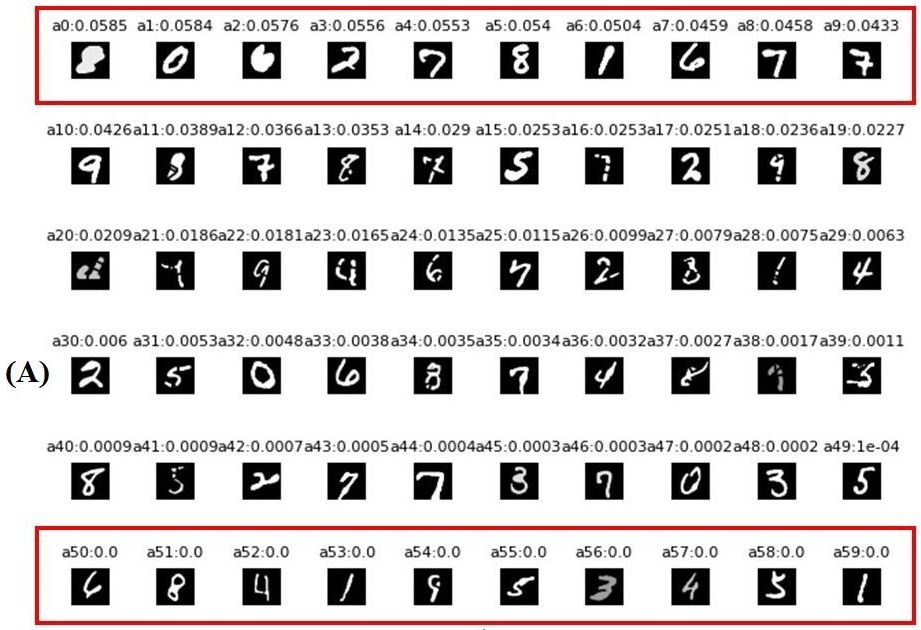}
\caption{Example of positive bags with their attention weights. Higher weights are assigned to thick digits (as shown in first row) which represent disease categories, diffuse and focal. Whereas lower attention weights correspond to plain or fragmented digits (as shown in the last row) which represent healthy and inactive.} \label{Sfig2}
\end{figure}

\begin{figure}[h]
\centering
\includegraphics[width=1.00\textwidth]{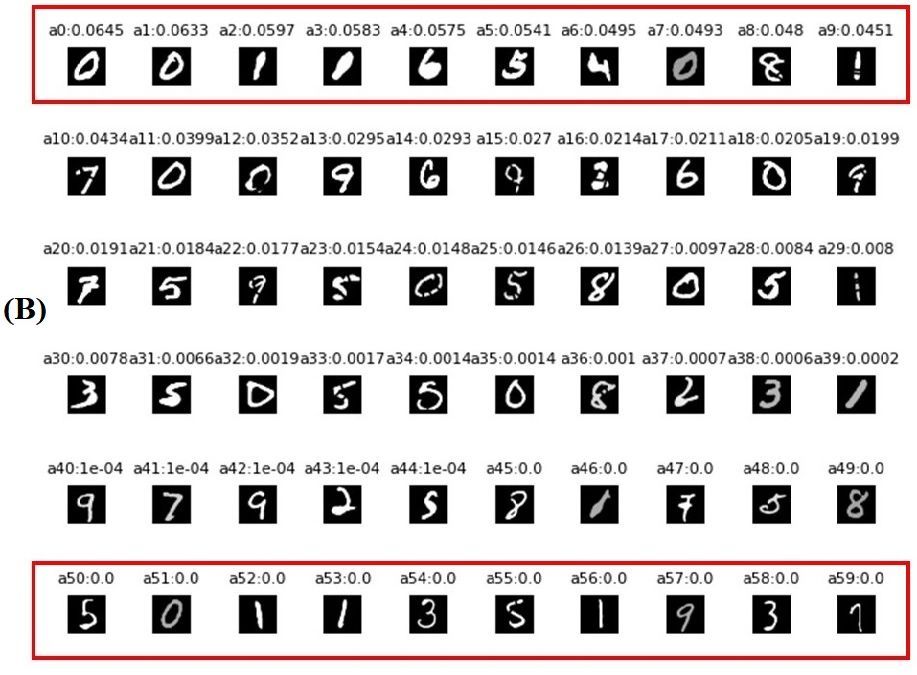}
\caption{Another example of positive bags with their attention weights (similar to Fig.~\ref{Sfig2}).} \label{Sfig3}
\end{figure}

\end{document}